\documentclass[letterpaper, 10 pt, conference]{ieeeconf}
\IEEEoverridecommandlockouts
\usepackage[usenames,dvipsnames]{xcolor}
\usepackage{graphics} %
\usepackage{graphicx, epstopdf}
\usepackage{amsmath,mathtools} %

\usepackage{amssymb}  %
\usepackage[nolist]{acronym}
\usepackage[hidelinks]{hyperref}
\usepackage{balance}
\usepackage[all]{hypcap} %
\usepackage[capitalize]{cleveref}
\usepackage[sort,compress]{cite}
\usepackage[normalem]{ulem}

\title{\LARGE \bf
Touch the Wind: Simultaneous Airflow, Drag and Interaction Sensing on a Multirotor
}

\author{Andrea Tagliabue$^{1,*}$, Aleix Paris$^{1,*}$, Suhan Kim$^{2}$, Regan Kubicek$^{2}$, Sarah Bergbreiter$^{2}$, Jonathan P. How$^{1}$%
\thanks{*equal contribution}%
\thanks{$^{1}$ Andrea Tagliabue, Aleix Paris, and Jonathan P. How are with the Department of Aeronautics and Astronautics, Massachusetts Institute of Technology. \tt\{atagliab, aleix, jhow\}@mit.edu}
\thanks{$^{2}$ Suhan Kim, Regan Kubicek, and Sarah Bergbreiter are with the Department of Mechanical Engineering, Carnegie Mellon University. \tt\{suhank, rkubicek, sbergbre\}@andrew.cmu.edu}%
}

\begin{document}

\maketitle
\thispagestyle{empty}
\pagestyle{empty}

\newcommand{\norm}[1]{\left\lVert#1\right\rVert}
\newcommand{\mnorm}[1]{\left\| #1 \right\|}

\begin{abstract}
Disturbance estimation for \acp{MAV} is crucial for robustness and safety. In this paper, we use novel, bio-inspired airflow sensors to measure the airflow acting on a \ac{MAV}, and we fuse this information in an \ac{UKF} to simultaneously estimate the three-dimensional wind vector, the drag force, and other interaction forces (e.g. due to collisions, interaction with a human) acting on the robot.
To this end, we present and compare a fully model-based and a deep learning-based strategy. The model-based approach considers the \ac{MAV} and airflow sensor dynamics and its interaction with the wind, while the deep learning-based strategy uses a \ac{LSTM} to obtain an estimate of the relative airflow, which is then fused in the proposed filter. We validate our methods in hardware experiments, showing that we can accurately estimate relative airflow of up to 4 m/s, and we can differentiate drag and interaction force.
\end{abstract}

\section{INTRODUCTION}
\label{sec:intro}

The deployment of \acp{MAV} in uncertain and constantly changing atmospheric conditions \cite{ZiplineV14:online,Flyabili5:online,Skydio2T39:online} requires the ability to estimate and adapt to disturbances such as the aerodynamic drag force applied by wind gusts. Simultaneously, as many new interaction-based missions \cite{STTRNavy83:online,VoliroAi75:online,tagliabue2019robust} arise, so increases the need to better differentiate between forces caused by aerodynamic disturbances and other sources of interaction \cite{augugliaro2013admittance, tagliabue2017collaborative, lew2019contact,paris2019dynamic}. Differentiating between aerodynamic drag force and interaction force can be extremely important for safety reasons. For example, the controller of a robot should react differently depending on whether a large disturbance is caused by a wind gust, or by a human trying to interact with the machine \cite{Myfinger52:online}.
\par
Distinguishing between drag and interaction disturbances can be challenging, as they both apply forces to the \ac{CoM} of the multirotor that cannot be easily differentiated by examining the inertial information
commonly available from the robot's onboard IMU or odometry estimator. Successful approaches for this task include a model-based method that measures the change in thrust-to-power ratio of the propellers caused by wind \cite{tomic2016flying} and an approach which monitors the frequency component of the  total disturbance (estimated via inertial information) to distinguish between the two possible sources of force \cite{tomic2015simultaneous}.

\begin{figure}[t]
    \centering
    \includegraphics[trim=0 350 0 200, clip, width=\columnwidth]{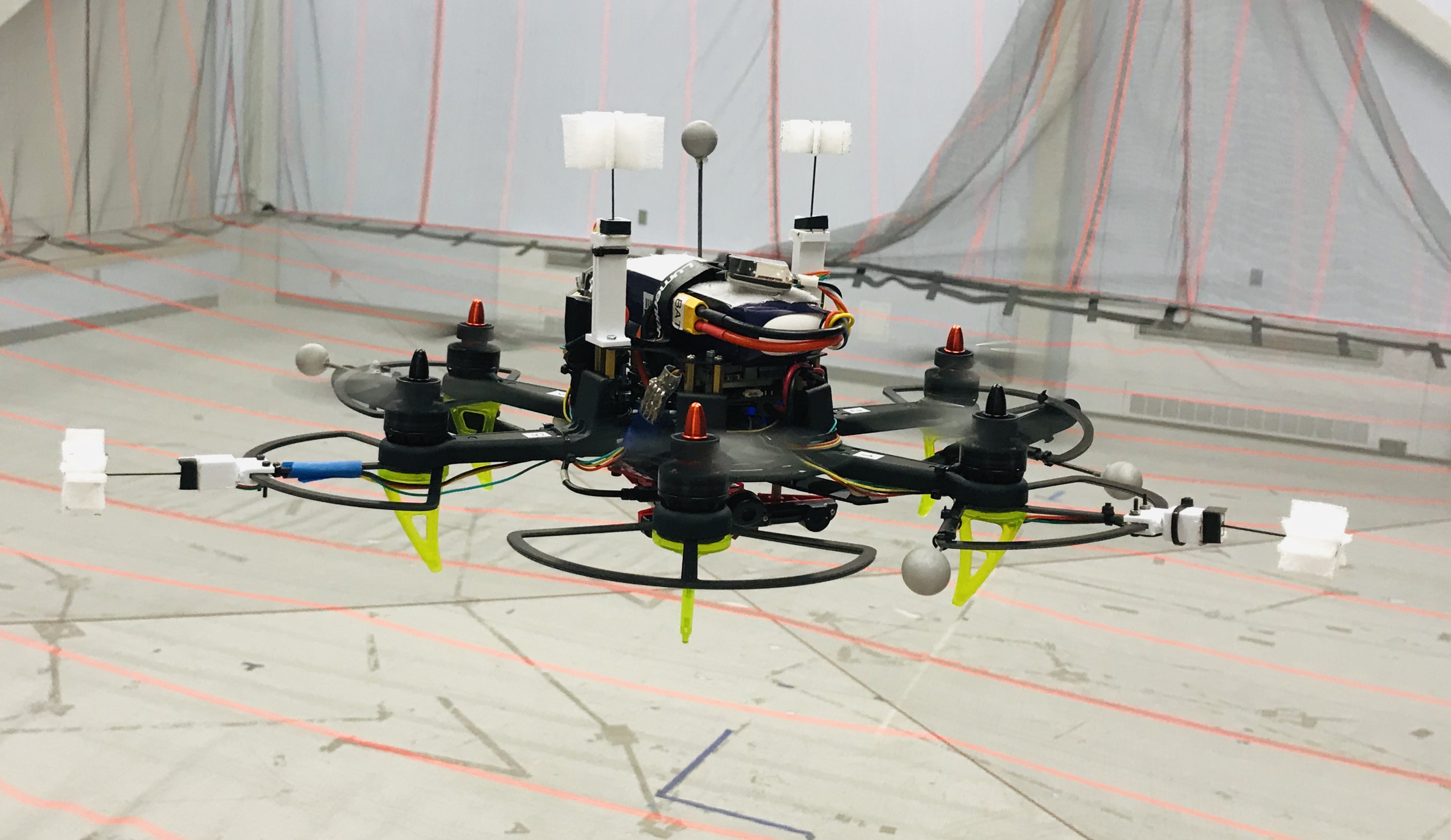}
    \caption{\ac{MAV} equipped with four bio-inspired airflow sensors used to estimate a three-dimensional wind vector from which we can distinguish aerodynamic drag from other forces (e.g., due to interaction).}
    \label{fig:multirotor_with_whiskers}
\end{figure}

This work presents a strategy for simultaneously estimating the interaction force and the aerodynamic drag disturbances using novel bio-inspired, whisker-like sensors that measure the airflow around a multirotor, as shown in \cref{fig:multirotor_with_whiskers}.
Our approach takes inspiration from the way insects sense airflow \cite{sane2007antennal}, which is by measuring the deflections caused by the aerodynamic drag force acting on the appendix of some receptors. By fusing the information of four heterogeneous airflow-sensors distributed across the surface of the robot, we can create a three-dimensional estimate of the relative velocity of the \ac{MAV} with respect to the surrounding airflow.
This information is then fused in a \ac{UKF}-based force estimator that uses an aerodynamic model together with the robot's pose and velocity to predict the wind, the expected drag force, and other interaction forces.

To account for the complex aerodynamic interactions between sensors and propellers \cite{prudden2018measuring,ventura2018high}, we extend this model-based approach (based on first-order physical principles) with a data-driven strategy. This strategy employs a \ac{RNN} based on a \ac{LSTM} network to provide an estimate of the relative airflow of the robot, which is then fused in the proposed model-based estimation scheme. We experimentally show that our approach achieves an accurate estimate of the relative airflow with respect to the robot with velocities up to $4$~m/s, and enables interaction forces and aerodynamic drag forces to be distinguished. We experimentally compare the model-based and learning-based approaches, highlighting their advantages and disadvantages.

To summarize, the contributions of this paper are: 
1) model- and deep learning-based strategies to simultaneously estimate wind, drag force, and other interaction forces using novel bio-inspired sensors similar to the one discussed in \cite{kim2019magnetically}; and 2) experimental validation of our approaches, showing that we can accurately estimate relative airflow of up to $4$~m/s and distinguish between interaction force and aerodynamic force.

\section{RELATED WORK}
Distinguishing between interaction and aerodynamic disturbance is a challenging task, and most of the current approaches focus on the estimation of one or the other disturbance. \textbf{Aerodynamic disturbances}: Accurate wind or airflow sensing is at the heart of the techniques employed for aerodynamic disturbance estimation. A common strategy is based on directly measuring the airflow surrounding the robot via sensors, such as pressure sensors \cite{bruschi2016wind}, ultrasonic sensors \cite{hollenbeck2018wind}, or whisker-like sensors \cite{deer2019lightweight}. Other strategies estimate the airflow via its inertial effects on the robot, for example using model-based approaches \cite{demitrit2017model, sikkel2016novel}, learning-based approaches \cite{shi2019neural, allison2019estimating}, or hybrid (model-based and learning-based) solutions \cite{marton2019hybrid}. \textbf{Generic wrench-like disturbances}: Multiple related works focus instead on estimating wrench disturbances, without explicitly differentiating for the effects of the drag force due to wind: \cite{augugliaro2013admittance, mckinnon2016unscented, tagliabue2019robust, tagliabue2017collaborative} propose a model-based approach which utilizes an \ac{UKF} for wrench estimation, while \cite{nisar2019vimo} proposes a factor graph-based estimation scheme.

\section{SENSOR DESIGN}
\label{sec:sensor_design}

\begin{figure}
    \centering
    \includegraphics[width=\linewidth]{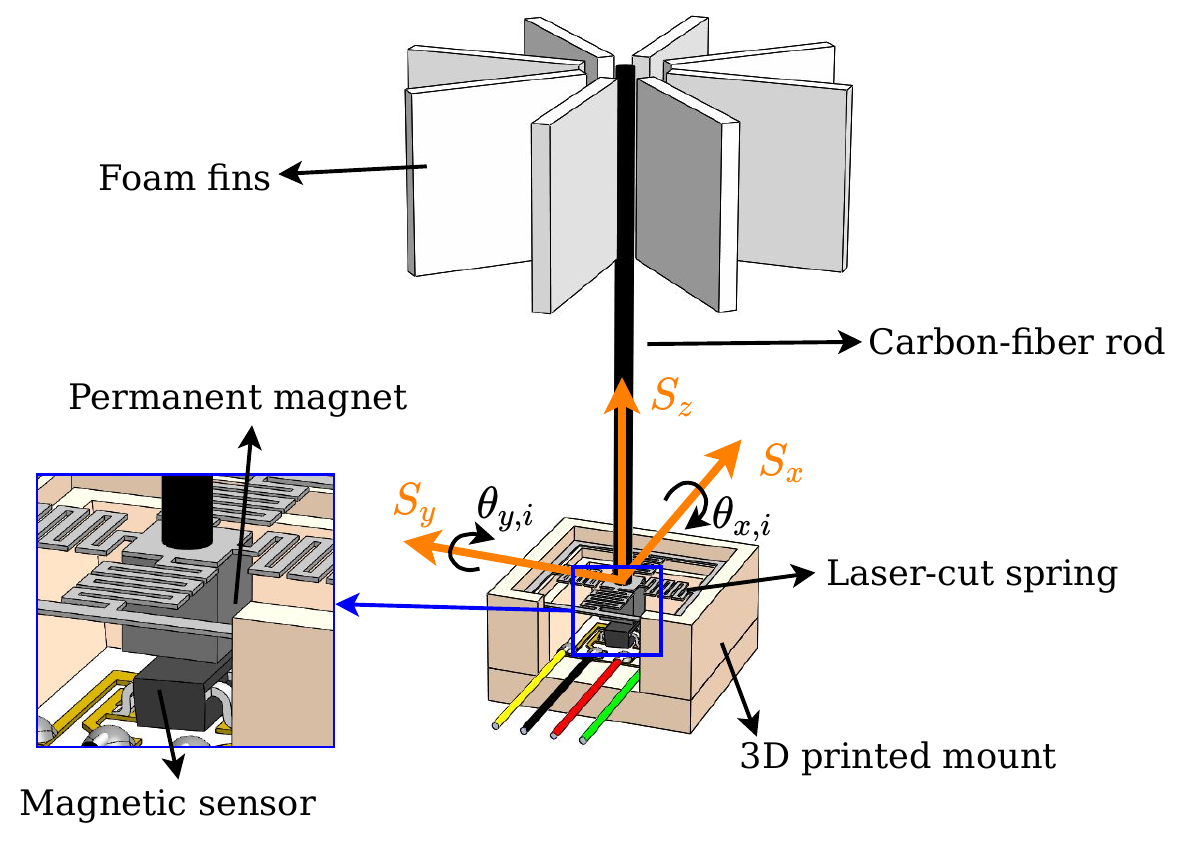}
    \caption{Illustration of an airflow sensor and its reference frame S, with the main components labeled.}
    \label{fig:sensor_model}
\end{figure}

In order to measure the relative wind in 3D affecting a MAV, lightweight, economical, and multi-directional sensors need to be used. This paper adopts sensors based in \cite{kim2019magnetically}, which satisfy these requirements.

\subsection{Sensor design and considerations}
The sensors, shown in \cref{fig:sensor_model}, consist of a base and an easily-exchangeable tip. The base is composed of a magnetic field sensor connected to a conditioning circuit that interfaces with the robot via I2C and a 3D-printed case that encloses the sensor.
The tip consists of a planar spring mounted in a 3D-printed enclosure that fits with the base, with a permanent magnet attached to its bottom and a carbon-fiber rod glued on the spring's top.
Eight foam fins are attached on the other end of this rod.
When the sensor is subjected to airflow, the drag force from the air on the fins causes a rotation about the center of the planar spring which results in a displacement of the magnet.
This displacement is then measured by the magnetic sensor.
The fins are placed with even angular distribution in order to achieve homogeneous drag for different airflow directions.
Foam and carbon fiber were chosen as the material of the fin structure due to their low density, which is crucial to minimize the inertia of the sensor. See \cite{kim2019magnetically} for more information about the sensor characteristics and manufacturing procedure.

Due to the complex aerodynamic interactions between the relative airflow and the blade rotor wakes, the sensor placement needs to be chosen carefully \cite{prudden2018measuring, ventura2018high}. To determine the best locations, we attached short pieces of string both directly on the vehicle and on metal rods extending away from it horizontally and vertically. We then flew the hexarotor indoors and observed that the pieces of string on top of the vehicle and on the propeller guards were mostly unaffected by the blade rotor wakes. Therefore, these are the two locations chosen to mount the sensors, as seen in \cref{fig:multirotor_with_whiskers}. They are distributed so that the relative airflow coming from any direction excites at least one sensor (that is, for at least one sensor, the relative airflow is not aligned with its length). %

\subsection{Sensor measurements}
The sensors detect the magnetic field $\mathbf{b} = (b_x, b_y, b_z)$, but the model outlined in \cref{subsec:sensor_model} requires the deflection angles of the $i$th sensor $\theta_{x,i}$ and $\theta_{y,i}$, which correspond to the rotation of the carbon fiber rod about the $x$ and $y$ axes in reference frame $S_i$. At the spring's equilibrium, the rod is straight and $\mathbf{b} = (0, 0, b_z)$, where $b_z > 0$ if the magnet's north pole is facing the carbon-fiber rod. The angles are then
\begin{equation}
    \boldsymbol{\theta}_i = 
    \begin{bmatrix}
    \theta_{x,i} \\ 
    \theta_{y,i} \\
    \end{bmatrix}
    =
    \begin{bmatrix}
    -\arctan{(b_y/b_z)} \\ 
    \hphantom{-}\arctan{(b_x/b_z)} \\
    \end{bmatrix}
    \label{eq:from_b_to_theta}
\end{equation}
Note that if the magnet was assembled with the south pole facing upward instead, $-\mathbf{b}$ must be used in \cref{eq:from_b_to_theta}.

\section{MODEL-BASED APPROACH}
\label{sec:model_based}

\begin{figure}
    \vspace*{.1in}
    \centering
    \includegraphics[width=1\columnwidth]{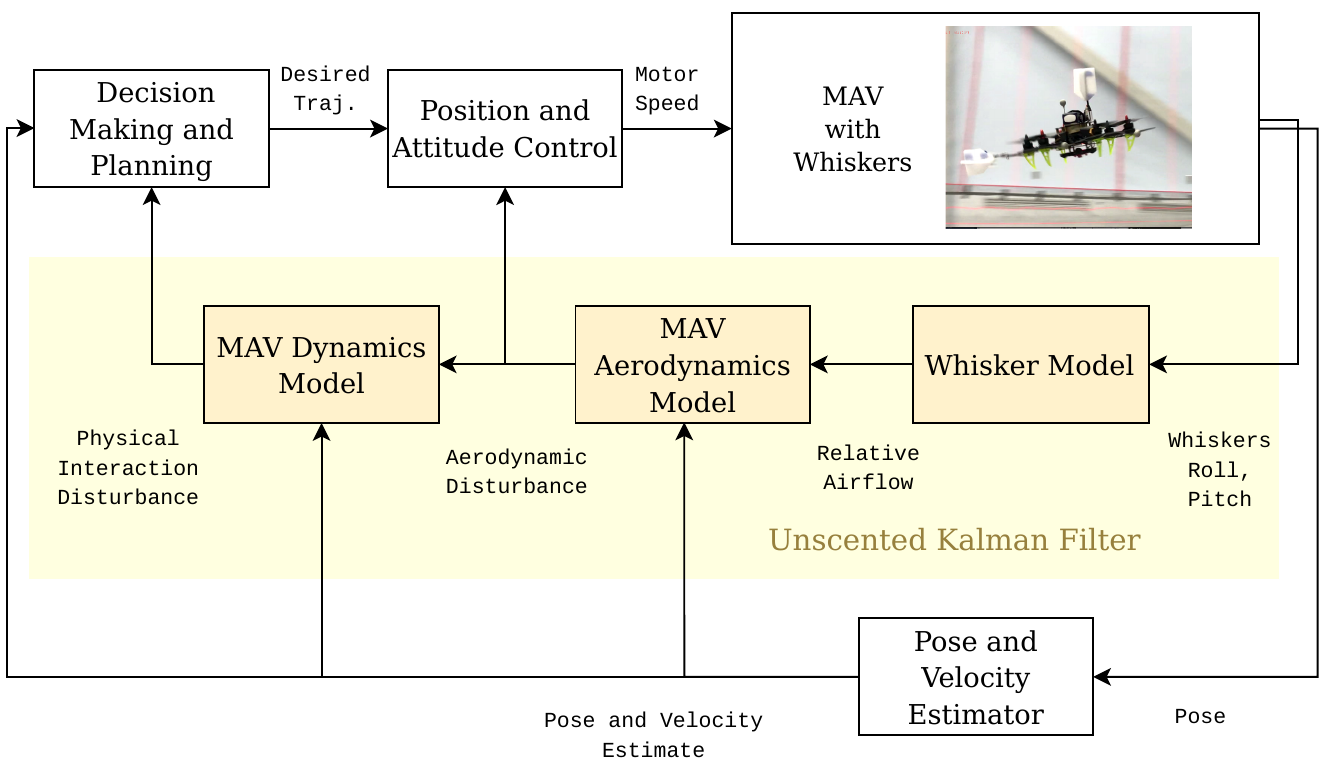}
    \caption{Diagram of the most important signals used by each step of the proposed model-based approach for simultaneous estimation of wind, drag force, and interaction force.}
    \label{fig:model_based}
\end{figure}

In this section, we present the model-based approach used to simultaneously estimate airflow, interaction force, and aerodynamic drag force on a MAV. The estimation scheme is based on the \ac{UKF} \cite{simon2006optimal} approach presented in our previous work \cite{tagliabue2019robust, tagliabue2017collaborative}, augmented with the ability to estimate a three-dimensional wind vector via the relative airflow measurements provided by the whiskers.
Here we summarize the approach and present a measurement model for the airflow sensors. A diagram of the most important signals and system-level blocks related to our approach is included in \cref{fig:model_based}.
\paragraph{Reference frame definition}
We consider an inertial reference frame W, a body-fixed reference frame B attached to the \ac{CoM} of the robot, and the $i$-th sensor reference frame $S_i$, with $i = 1,\dots,N$, as shown in \cref{fig:sensor_model}. 

\subsection{\ac{MAV} dynamic model} 
We consider a \ac{MAV} of mass $m$ and inertia tensor $\mathbf{J}$, and the dynamic equations of the robot can be written as
\begin{equation}
    \begin{split}
        \prescript{}{W}{\dot{\mathbf{p}}} = & \prescript{}{W}{\mathbf{v}} \\
        \dot{\mathbf{R}}_{W}^{B} = & \mathbf{R}_{W}^{B}[\prescript{}{B}{\boldsymbol{\omega}}\times] \\
        m\prescript{}{W}{\dot{\mathbf{v}}} = &  \mathbf{R}_{W}^{B} \prescript{}{B}{\mathbf{f}}_{\text{cmd}} + \prescript{}{W}{\mathbf{f}_\text{drag}} + m \prescript{}{W}{\mathbf{g}} + \prescript{}{W}{\mathbf{f}}_\text{touch} \\
        \mathbf{J} \prescript{}{B}{\dot{\boldsymbol{\omega}}} = & -\prescript{}{B}{\boldsymbol{\omega}} \times \mathbf{J} \prescript{}{B}{\boldsymbol{\omega}} + \prescript{}{B}{\boldsymbol{\tau}}_{\text{cmd}} \\
    \end{split}
    \label{eq:mav_dynamic_model}
\end{equation}
where $\mathbf{p}$ and $\mathbf{v}$ represent the position and velocity of the MAV, respectively, $\mathbf{R}_{W}^{B}$ is the rotation matrix representing the attitude of the robot (i.e., such that a vector $\prescript{}{W}{\mathbf{p}} = \mathbf{R}_{W}^{B} \prescript{}{B}{\mathbf{p}}$), and $[\times]$ denotes the skew-symmetric matrix.
The vector $\prescript{}{B}{\mathbf{f}}_{\text{cmd}} = \prescript{}{B}{\mathbf{e}}_3 f_{\text{cmd}}$ is the thrust force produced by the propellers along the $z$-axis of the body frame, $\prescript{}{W}{\mathbf{g}} = -\prescript{}{W}{\mathbf{e}}_3 g$ is the gravitational acceleration, and $\prescript{}{W}{\mathbf{f}}_\text{touch}$ is the interaction force expressed in the inertial frame.
For simplicity we have assumed that interaction and aerodynamic disturbances do not cause any torque on the \ac{MAV}, due to its symmetric shape and the fact that interactions (in our hardware setup) can only safely happen in proximity of the center of mass of the robot.
Vector $\prescript{}{B}{\boldsymbol{\tau}}_{\text{cmd}}$ represents the torque generated by the propellers and $\prescript{}{B}{\boldsymbol{\omega}}$ the angular velocity of the MAV, both expressed in the body reference frame.
Here $\mathbf{f}_\text{drag}$ is the aerodynamic drag force on the robot, expressed as an isotropic drag \cite{tagliabue2019model}
\begin{equation}
\begin{split}
\prescript{}{W}{\mathbf{f}}_\text{drag} = & (\mu_1 \text{v}_\infty + \mu_2 \text{v}_\infty^2)\prescript{}{W}{\mathbf{e}}_{\text{v}_\infty} = f_\text{drag} \prescript{}{W}{\mathbf{e}}_{\text{v}_{\infty}} \\ 
 & _W{\mathbf{e}}_{\text{v}_{\infty}}  = \frac{\prescript{}{W}{\mathbf{v}}_\infty}{\text{v}_\infty}, \quad \text{where} \quad \text{v}_\infty = \mnorm{\prescript{}{W}{\mathbf{v}}_\infty},
\label{eq:drag_force}
\end{split}
\end{equation}
$\prescript{}{W}{\mathbf{v}}_\infty$ is the velocity vector of the relative airflow acting on the \ac{CoM} of the \ac{MAV} (expressed in the inertial frame) 
\begin{equation}
   \prescript{}{W}{\mathbf{v}}_\infty = \prescript{}{W}{\mathbf{v}}_\text{wind} - \prescript{}{W}{\mathbf{v}},
   \label{eq:relative_airflow}
\end{equation}
and $\prescript{}{W}{\mathbf{v}}_\text{wind}$ is the velocity vector of the wind expressed in the inertial frame. 

\subsection{Airflow sensor model}\label{subsec:sensor_model}
We consider the $i$-th airflow sensor to be rigidly attached to the body reference frame $B$, with $i=1,\ldots,N$. The reference frame of each sensor is translated with respect to $B$ by a vector $\prescript{}{B}{\mathbf{r}}_{S_i}$ and rotated according to the rotation matrix $\mathbf{R}_{S_i}^B$. 
To derive a model of the whiskers subject to aerodynamic drag, we make the following assumptions. Each whisker is massless; its tilt angle is not significantly influenced by the accelerations from the base $B$ (due to the high stiffness of its spring and the low mass of the fins), but is subject to the aerodynamic drag force $\mathbf{f}_{\text{drag}, i}$. 

We further assume that each sensor can be modeled as a stick hinged at the base via a linear torsional spring. Each sensor outputs the displacement angle $\theta_{x,i}$ and $\theta_{y,i}$, which correspond to the rotation of the stick around the $x$ and $y$ axis of the $S_i$ reference frame. 
We can then express the aerodynamic drag force acting on the aerodynamic surface of each sensor $\prescript{}{S_i}{\mathbf{f}}_{\text{drag}, i}$ as a function of the (small) displacement of the angle
\begin{equation}
    \mathbf{S}_{x,y}\prescript{}{S_i}{\mathbf{f}}_{\text{drag}, i} \approx 
    \begin{bmatrix}
    0 & l_i k_i \\
    - l_i k_i & 0 \\
    \end{bmatrix}
    \begin{bmatrix}
    \theta_{x,i} \\ 
    \theta_{y,i} \\
    \end{bmatrix}
    = \mathbf{K}_i \boldsymbol{\theta}_i
    \label{eq:model:drag_and_spring}
\end{equation}
where $k_i$ represents the stiffness of the torsional spring, $l_i$ the length of the sensor, and 
\begin{equation}
    \mathbf{S}_{x,y} = 
    \begin{bmatrix}
    1 & 0 & 0 \\
    0 & 1 & 0 \\
    \end{bmatrix}
\end{equation}
captures 
the assumption that the aerodynamic drag acting on the $z$-axis of the sensor is small (given the fin shapes) and has a negligible effect on the sensor deflection.

We now consider the aerodynamic force acting on a whisker. Assuming a non-isotropic drag, proportional to the squared relative velocity w.r.t. the relative airflow, we obtain
\begin{equation}
\begin{split}
    \prescript{}{S_i}{\mathbf{f}}_{\text{drag}, i} = &\frac{\rho}{2}c_{D,i}\mathbf{A}_i
    \norm{\prescript{}{S_i}{\mathbf{v}}_{\infty,i}}\prescript{}{S_i}{\mathbf{v}}_{\infty,i} 
\end{split}
\label{eq:model:whisker_drag}
\end{equation}
where $\rho$ is the density of the air, $c_D$ is the aerodynamic drag coefficient, $\mathbf{A}_i = \text{diag}([a_{xy,i}, a_{xy,i}, a_{z}]^\top)$ is the aerodynamic section of each dimension, and $c_{D,i}$ the corresponding drag coefficient. Due to the small vertical surface of the fin of the sensor, we assume $a_z = 0$. The vector $\prescript{}{S_i}{\mathbf{v}}_{\infty,i}$ is the velocity of the relative airflow experienced by the $i$-th whisker, and expressed in the $i$-th whisker reference frame, and can be obtained as %
\begin{equation}
    \prescript{}{S_i}{\mathbf{v}}_{\infty,i} =  {\mathbf{R}_{B}^{S_i}}^\top(\prescript{}{B}{\mathbf{v}}_\infty
    - \prescript{}{B}{\boldsymbol{\omega}} \times \prescript{}{B}{\mathbf{r}}_{S_i}) \\
    \label{eq:rel_airflow_whisker_in_whisker_frame}
\end{equation}
where $\prescript{}{B}{\mathbf{v}}_{\infty}$ 
is the relative airflow in the \ac{CoM} of the robot expressed in the body frame, given by: 

\begin{equation}
   \prescript{}{B}{\mathbf{v}}_\infty = {\textbf{R}_{W}^{B}}^\top\prescript{}{W}{\mathbf{v}}_\infty =  {\textbf{R}_{W}^{B}}^\top(\prescript{}{W}{\mathbf{v}}_\text{wind} - \prescript{}{W}{\mathbf{v}}) 
   \label{eq:rel_airflow_in_body}
\end{equation}

\subsection{Model-based estimation scheme}

\subsubsection{Process model, state and output}
We discretize the \ac{MAV} dynamic model described in \cref{eq:mav_dynamic_model} augmenting the state vector with the unknown wind $\prescript{}{W}{\mathbf{v}}_{\text{wind,k}}$ and unknown interaction force $\prescript{}{W}{\mathbf{f}}_{\text{touch},k}$ that are to be estimated. We assume that these two state variables evolve as:
\begin{equation}
\begin{split}
     \prescript{}{W}{\mathbf{f}}_{\text{touch},k+1} = \prescript{}{W}{\mathbf{f}}_{\text{touch},k} + \boldsymbol{\epsilon}_{{f},k} \\
     \prescript{}{W}{\mathbf{v}}_{\text{wind,k+1}} = \prescript{}{W}{\mathbf{v}}_{\text{wind,k}} + \boldsymbol{\epsilon}_{{v},k} \\
\end{split}
\end{equation}
where $\boldsymbol{\epsilon}_{f,k}$ and $\boldsymbol{\epsilon}_{v,k}$ represent the white Gaussian process noise, with covariances used as tuning parameters. 

The full, discrete time state of the system used for estimation is
\begin{equation}
\begin{aligned}
    {{\boldsymbol{x}}_k}^\top \! =  \{ &
        {\prescript{}{W}{\mathbf{p}}_k}^\top, 
        {\mathbf{q}_{W,k}^{B}}^\top, 
        {\prescript{}{W}{\mathbf{v}_k}}^\top, \\ &{\prescript{}{B}{\boldsymbol{\omega}}_k}^\top, {\prescript{}{W}{\mathbf{f}}_{\text{touch},k}}^\top, {\prescript{}{W}{\mathbf{v}}_{\text{wind},k}}^\top
    \}
\end{aligned}
\label{eq:ukf_state}
\end{equation}
where $\mathbf{q}_{W,k}^{B}$ is the more computationally efficient quaternion-based attitude representation of the robot, obtained from the rotation matrix $\mathbf{R}_{W,k}^{B}$.

The filter output is then 
\begin{equation}
    {\mathbf{y}_k}^\top  \! = \!
    \{
        {\prescript{}{W}{\mathbf{f}}_{\text{touch},k}}^\top, {\prescript{}{W}{\mathbf{v}}_{\text{wind},k}}^\top,
        {\prescript{}{B}{\mathbf{v}}_{\infty,k}}^\top,
        {\prescript{}{W}{\mathbf{f}}_{\text{drag},k}}^\top
    \}
\end{equation}
where $\prescript{}{W}{\mathbf{f}}_{\text{drag},k}$ is obtained from \cref{eq:drag_force} and \cref{eq:relative_airflow}, and $\prescript{}{B}{\mathbf{v}}_{\infty,k}$ is obtained from \cref{eq:rel_airflow_in_body}.

\subsubsection{Measurements and measurement model}
We assume that two sets of measurements are available asynchronously:
\paragraph{Odometry}
The filter fuses odometry measurements (position $\prescript{}{}{\hat{\mathbf{p}}}_k$, attitude $\prescript{}{}{\hat{\mathbf{q}}}_{W,k}^{B}$, linear velocity $\prescript{}{W}{\hat{\mathbf{v}}}$ and angular velocity $\prescript{}{B}{\hat{\boldsymbol{\omega}}}$) provided by a cascaded state estimator
\begin{equation}
\begin{split}
    {\mathbf{z}_{\text{odometry},k}}^\top & =
    \{
        {\prescript{}{W}{\hat{\mathbf{p}}}_k}^\top, {\prescript{}{}{\hat{\mathbf{q}}}_{W,k}^{B}}^\top, {\prescript{}{W}{\hat{\mathbf{v}}}}^\top, {\prescript{}{B}{\hat{\boldsymbol{\omega}}}}^\top
    \} \\
\end{split}
\end{equation}
the odometry measurement model is linear, as shown in \cite{tagliabue2019robust}.
\paragraph{Airflow sensors}
We assume that the $N$ sensors are sampled synchronously, providing the measurement vector
\begin{equation}
    {\mathbf{z}_{\text{airflowsensor},k}}^\top  = 
    \{ 
        {\hat{\boldsymbol{\theta}}_{1,k}}^\top,\ldots,{\hat{\boldsymbol{\theta}}_{N,k}}^\top \} = {\hat{\boldsymbol{\theta}}_k}^\top \\
\end{equation}
The associated measurement model for the $i$-th sensor can be obtained by combining \cref{eq:model:drag_and_spring} and \cref{eq:model:whisker_drag}
\begin{equation}
    \boldsymbol{\theta}_{i,k} = 
    \frac{\rho}{2}c_D 
    {\mathbf{K}_i}^{-1}
    \mathbf{S}_{x,y}
    \mathbf{A}_i
    \norm{\prescript{}{S_i}{\mathbf{v}}_{\infty,i,k}}\prescript{}{S_i}{\mathbf{v}}_{\infty,i,k}  
    \label{eq:whisker_measurement}
\end{equation} where $\prescript{}{S_i}{\mathbf{v}}_{\infty,i,k}$ is obtained using information about the attitude of the robot $\mathbf{q}_{W,k}^{B}$, its velocity $\prescript{}{W}{\mathbf{v}_k}$, and angular velocity $\prescript{}{B}{\boldsymbol{\omega}_k}$, and the estimated windspeed $\prescript{}{W}{\mathbf{v}}_{\text{wind}}$ as described in \cref{eq:rel_airflow_whisker_in_whisker_frame} and \cref{eq:rel_airflow_in_body}. The synchronous measurement update is obtained by repeating \cref{eq:whisker_measurement} for every sensor $i=1,\ldots,N$.

\subsubsection{Prediction and update step}
\paragraph{Prediction}
The prediction step (producing the \textit{a priori} state estimate) \cite{simon2006optimal} is performed using the \ac{USQUE} \cite{crassidis2003unscented} prediction technique for the attitude quaternion. The process model is propagated using the commanded thrust force $f_{\text{cmd}}$ and torque $\prescript{}{B}{\boldsymbol{\tau}}_{\text{cmd}}$ output of the position and attitude controller on the \ac{MAV}.
\paragraph{Update}
The odometry measurement update step is performed using the linear Kalman filter update step \cite{simon2006optimal}, while the airflow-sensor measurement update is performed via the Unscented Transformation \cite{simon2006optimal} due to the non-linearities in the associated measurement model.

\section{DEEP-LEARNING BASED APPROACH}
\label{sec:learning_based}

\begin{figure}
    \vspace*{.05in}
    \centering
    \includegraphics[width=\linewidth]{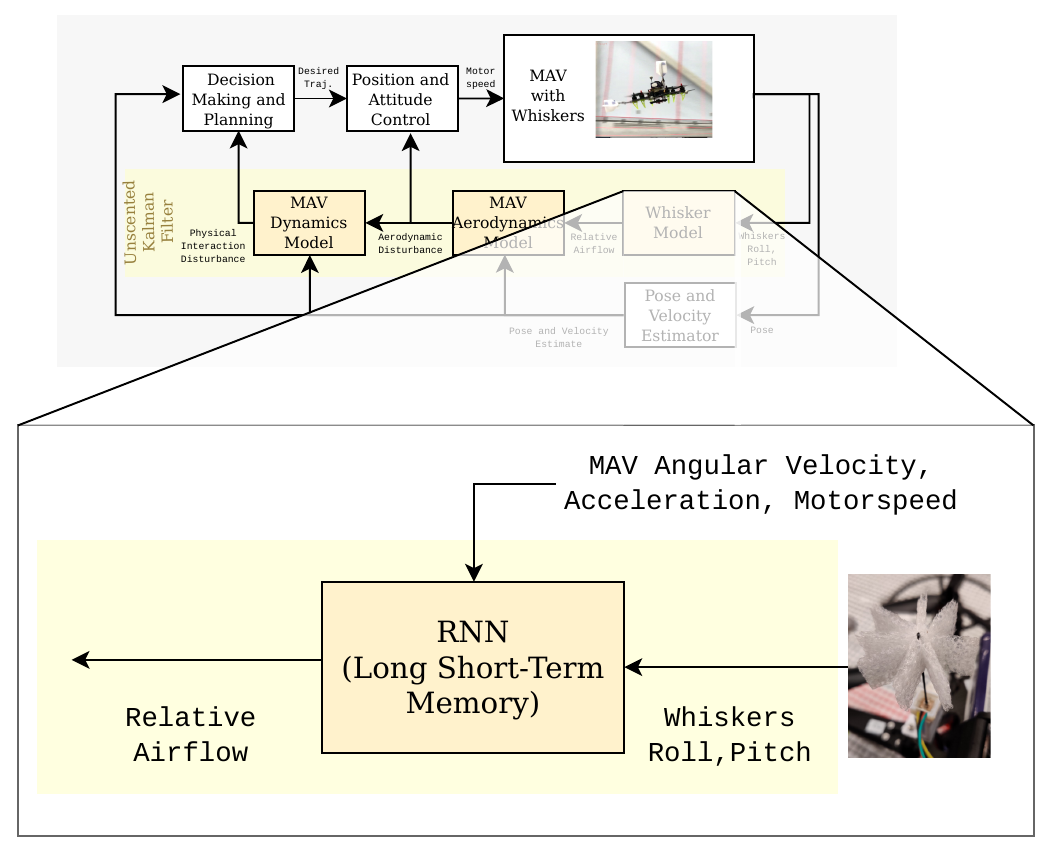}
    \caption{Signal diagram of the interface between the learning-based and the data-driven approach.}
    \label{fig:data_driven}
\end{figure}
In this section we present a deep-learning based strategy, which makes use of a \ac{RNN} based on the \ac{LSTM} architecture to create an estimate of the relative airflow $\prescript{}{B}{\mathbf{v}}_{\infty}$ using the airflow sensors and other measurements available on board of the robot.
The complexity in modeling the effects of the aerodynamic interference caused by the airflow between the propellers, the body of the robot and the surrounding air, as observed in the literature \cite{prudden2018measuring} \cite{ventura2018high} and in our own experimental results, motivates the use of a learning-based strategy to map sensors' measurement to relative airflow.

\subsection{Output and inputs}
The output of the network is the relative airflow $\prescript{}{B}{\mathbf{v}}_{\infty}$ of the \ac{MAV}.
The inputs to the network are the airflow sensor measurements $\boldsymbol{\theta}$, the angular velocity of the robot $\prescript{}{B}{\boldsymbol{\omega}}$, the raw acceleration measurement from the IMU and the normalized throttle commanded to the six propellers (which ranges between 0 and 1).
The sign of the throttle is changed for the propellers spinning counterclockwise, in order to provide information to the network about the spinning direction of each propeller.
The reason for the choice of the input is dictated by the derivation from our model-based approach: from \cref{eq:rel_airflow_whisker_in_whisker_frame} and \cref{eq:model:whisker_drag} we observe that the relative airflow depends on the angle of the sensors and on the angular velocity of the robot.
The acceleration from the IMU is included to provide information about hard to model effects, such as the orientation of the body frame w.r.t. gravity (which causes small changes in the angle measured by the sensors), as well as the effects of accelerations of the robot.
Information about the throttle and spinning direction of the propellers is instead added to try to capture the complex aerodynamic interactions caused by their induced velocity.
We chose to express every output and input of the network in the body reference frame, in order to make the network invariant to the orientation of the robot, thus potentially reducing the amount of training data needed. 

\subsection{Network architecture}
We employ an \ac{LSTM} architecture, which is able to capture time-dependent effects \cite{lipton2015critical, goodfellow2016deep}, such as, in our case, the dynamics of the airflow surrounding the robot and the dynamics of the sensor. We chose 
a 2-layer LSTM, with the size of the hidden layer set to 16 (with the input size, 20, and the output size, 3). We add a single fully connected layer to the output of the network, mapping the hidden layer into the the desired output size. 

\subsection{Interface with the model-based approach}
The \ac{UKF} treats the \ac{LSTM} output as a new sensor which provides relative airflow measurements $\prescript{}{B}{\hat{\mathbf{v}}}_\infty$, replacing the airflow sensor's measurement model provided in \cref{sec:model_based}. The output of the LSTM is fused via the measurement model in \cref{eq:rel_airflow_in_body}, using the Unscented Transformation.
A block-diagram representing the interface between learning-based approach and model-based approach is represented in \cref{fig:data_driven}.

\section{EXPERIMENTAL EVALUATION}
\label{sec:results}
\subsection{System identification}\label{subsec:sysid}
\subsubsection{Drag force}
Estimating the drag force acting on the vehicle is required to differentiate from force due to relative airflow and force due to other interactions with the environment. To this purpose, the vehicle was commanded to follow a circular trajectory at speeds of 1 to 5 m/s, keeping its altitude constant (see \cref{subsec:implementation} for more information about the trajectory generator). In this scenario, the thrust produced by the MAV's propellers $\hat{f}_{\text {thrust }}$ is
\begin{equation}
\hat{f}_{\text {thrust }}=\frac{m}{\cos \phi \cos \theta} g
\end{equation}
where $m$ is the vehicle's mass, $g$ is the gravity acceleration, and $\phi$ and $\theta$ are respectively the roll and pitch angles of the MAV. The drag force is then
\begin{equation}
    \hat{f}_{\text {drag }}=
     \left( \prescript{}{B}{\hat{\mathbf{f}}_{\text{thrust }}} - m\prescript{}{B}{\dot{\mathbf{v}}} \right) \cdot \prescript{}{B}{\mathbf{e}_{v}}
\end{equation}
where $\prescript{}{B}{\mathbf{\hat{f}_{\text {thrust }}}} = [0, 0, \hat{f}_{\text {thrust }}]$, and $\prescript{}{B}{\mathbf{e}_{v}}$ is the unit vector in the direction of the vehicle's velocity in body frame. By fitting a second-degree polynomial to the collected data, we obtain $\mu_1 = 0.20$ and $\mu_2 = 0.07$ (see \cref{eq:drag_force}).

\subsubsection{Sensor parameters identification}
\begin{figure}
    \centering
    \includegraphics[width=\linewidth, trim=0 0 0 0, clip,]{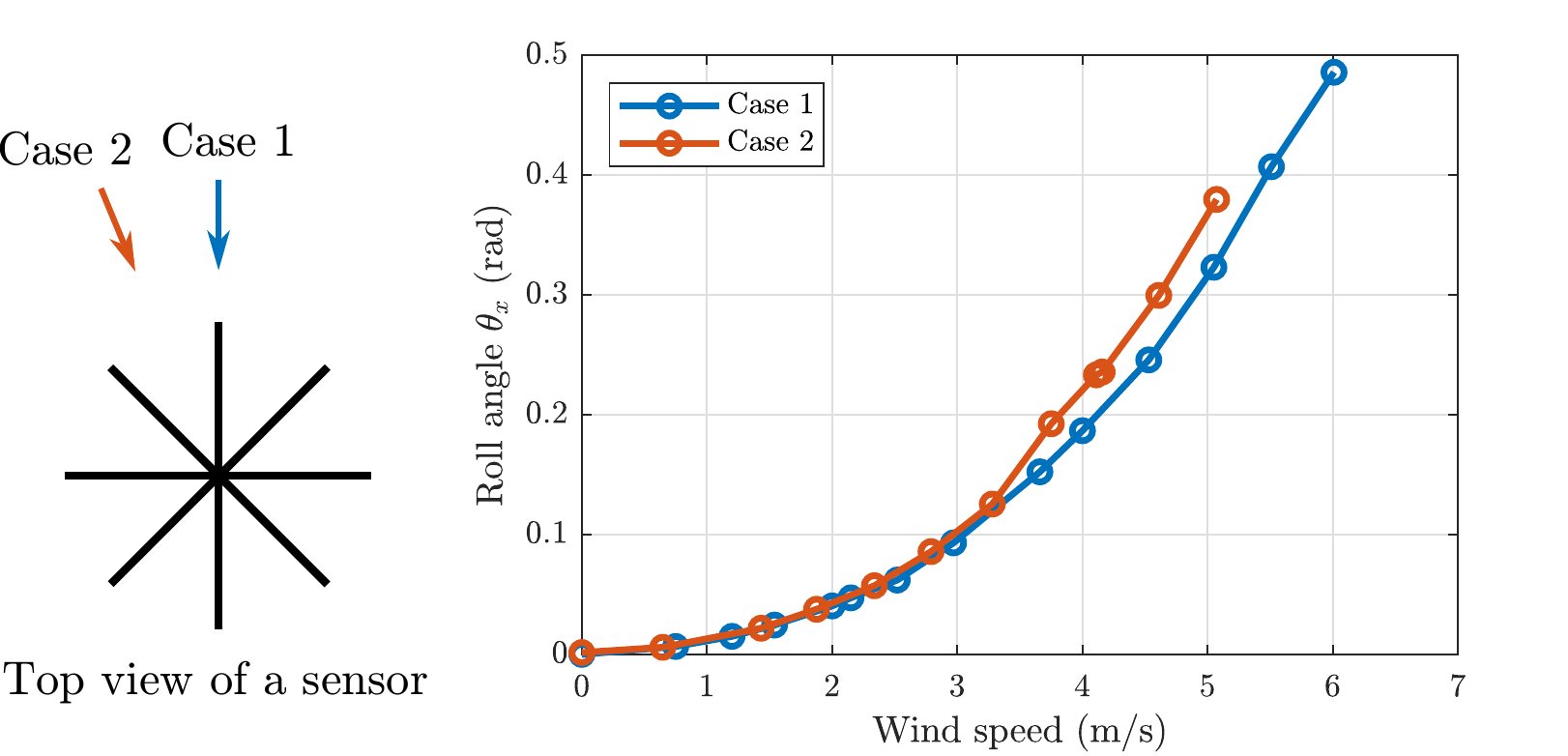}
    \caption{Roll deflection angle of the sensor as a function of the wind speed, for the case where the wind vector is aligned with a fin (1), and the case where it is most misaligned with a fin (2).}
    \label{fig:octaflower_comparison}
\end{figure}

The parameters required to fuse the output $\boldsymbol{\theta}_i$ of $i$-th airflow sensor are its position $\prescript{}{B}{\mathbf{r}}_{S_i}$ and rotation $\mathbf{R}_{B}^{S_i}$ with respect to the body frame B of the \ac{MAV}, and a lumped parameter coefficient $c_i$  mapping the relative airflow $\prescript{}{S_i}{\mathbf{v}}_{\infty,i}$  to the measured deflection angle $\boldsymbol{\theta}_i$. The coefficient $c_i = \frac{\rho}{2}c_{D_i}\frac{a_{xy_i}}{kl}$ can be obtained by re-arranging \cref{eq:whisker_measurement} and by solving 
\begin{align}
c_i = \frac{\norm{\boldsymbol{\theta}_i}}{
    \norm{\prescript{}{S_i}{\mathbf{v}_{\infty,i}}} \hspace{3pt}
    \norm{\begin{bsmallmatrix}
        0 & -1 & 0 \\
        1 & 0 & 0 \\
        \end{bsmallmatrix}
    \prescript{}{S_i}{\mathbf{v}}_{\infty,i} }}
\end{align}
and the velocity $\prescript{}{S_i}{\mathbf{v}}_{\infty,i}$ is obtained from indoor flight experiments (assuming no wind, so that $\prescript{}{W}{\mathbf{v}}_\infty = - \prescript{}{W}{\mathbf{v}}$), or by wind tunnel experiments. Wind tunnel experiments have also been used to validate our model choice (quadratic relationship between wind speed and sensor deflection), as show in \cref{fig:octaflower_comparison}. Furthermore, these experiments also confirmed our assumption on the structure of $\mathbf{A}_i$, i.e., the variation of the sensor's deflection with respect to the direction of the wind speed is small and therefore it can be considered that $a_x = a_y = a_{xy}$.

\subsubsection{LSTM training}
We train the \ac{LSTM} using two different datasets collected in indoor flight. In the first flight the hexarotor follows a circular trajectory at a set of constant velocities ranging from 1 to 5 m/s, spaced of 1 m/s each. In the second data-set we command the robot via a joystic, making aggressive maneuvers, while reaching velocities up to 5.5 m/s. 
Since the robot flies indoor (and thus wind can be considered to be zero) we assume that the relative airflow of the \ac{MAV} $\prescript{}{B}{\mathbf{v}_\infty}$ corresponds to its estimated velocity $-\prescript{}{B}{\mathbf{v}}$, which we use to train the network. The network is implemented and trained using PyTorch \cite{paszke2019pytorch}. The data is pre-process by re-sampling it at 50 Hz, since the inputs of the network used for training have different rates (e.g. 200 Hz for the acceleration data from the IMU and 50 Hz from the airflow sensors).
The network is trained for 400 epochs using sequences of 5 samples of length, with a learning rate of 10$^{-4}$, using the Adam optimizer \cite{kingma2014adam} and the \ac{MSE} loss. 
Unlike the model-based approach, the LSTM does not require any knowledge of the position and orientation of the sensors, nor the identification of the lumped parameter for each sensor. Once the network has been trained, however, it is not possible to reconfigure the position or the type of sensors used.

\subsection{Implementation details}\label{subsec:implementation}
\subsubsection{System architecture}
We use a custom-built hexarotor of 1.31 kg of mass. The pose of the robot is provided by a motion capture system, while odometry information is obtained by an estimator running on-board, which fuses the pose information with the inertial data from an IMU. Our algorithms run on the onboard Nvidia Jetson TX2 and are interfaced with the rest of the system via ROS. We use Aerospace Controls Laboratory's snap-stack \cite{acl_snap_stack} for controlling the \ac{MAV}.

\subsubsection{Sensor driver}
The sensors are connected via I2C to the TX2. A ROS node (sensor driver) reads the magnetic field data at 50~Hz and publishes the deflection angles as in \cref{eq:from_b_to_theta}. 
Slight manufacturing imperfections are handled via an initial calibration of offset angles . 
The sensor driver rejects any measured outliers by comparing each component of $\mathbf{b}$ with a low-pass filtered version. If the difference is large, the measurement is discarded, but the low-pass filter is updated nevertheless. Therefore, if the sensor deflects very rapidly and the measurement is incorrectly regarded an outlier, the low-pass filtered $\mathbf{b}$ quickly approaches the true value and consequent negative positives do not occur.

\begin{figure*}[ht]
\centering
\includegraphics[trim=0 0 0 0, clip, width=1.0\textwidth]{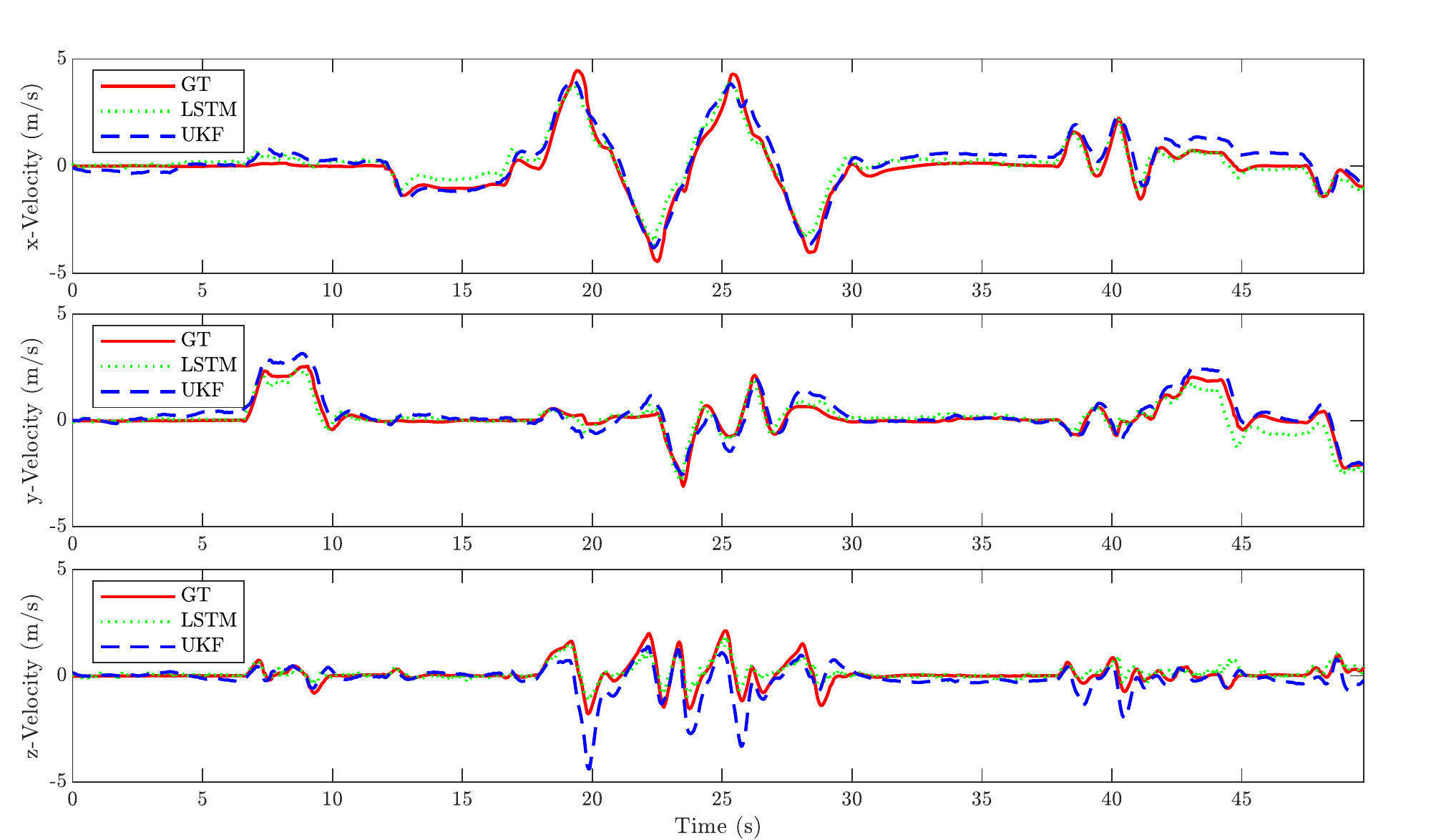}
\vspace*{-.15in}
\caption{Comparison of the relative velocity estimated by the model based (UKF) and  the learning-based (LSTM) approaches.
We assume that the ground truth (GT) is given by the velocity of the robot.}
\label{fig:rel_velocity}
\end{figure*}

\subsubsection{Trajectory generator}
A trajectory generator ROS node commands the vehicle to follow a circular path at different constant speeds or a line trajectory between two points with a maximum desired velocity. This node also handles the finite state machine transitions: take off, flight to the initial position of the trajectory, execution of the trajectory, and landing where the vehicle took off. We use this trajectory generator to identify the drag coefficient of the MAV (see \cref{subsec:sysid}), to collect data for training, and to execute the experiments described below.

\subsection{Relative airflow estimation}
For this experiment, we commanded the vehicle with a joystick along our flight space at different speeds, to show the ability of our approach to estimate the relative airflow. Since the space is indoors (no wind), we assume that the relative airflow is opposite to the velocity of the MAV. We thus compare the velocity of the MAV (obtained from a motion capture system) to the opposite relative airflow estimated via the model-based strategy and the deep-learning based strategy.

\Cref{fig:rel_velocity} shows the results of the experiment. Each subplot presents the velocity of the vehicle in body frame. The ground truth (GT) in red is the MAV's speed obtained via the motion capture system, the green dotted line represents the relative airflow velocity in body frame $-\prescript{}{B}{\mathbf{v}}_\infty$ as estimate via the deep-learning based strategy (LSTM), and the blue dashed line represents $-\prescript{}{B}{\mathbf{v}}_\infty$ as estimated by the the fully model-based strategy (UKF). 
The root mean squared errors of the UKF and LSTM's estimation for this experiment are shown in \cref{tab:rmse_rel_vel}. The results demonstrate that both approaches are effective, but show that the LSTM is more accurate.

\subsection{Wind gust estimation}
\begin{figure}
    \centering
    \includegraphics[width=1\columnwidth,trim={0cm 0cm 0cm 0cm},clip]{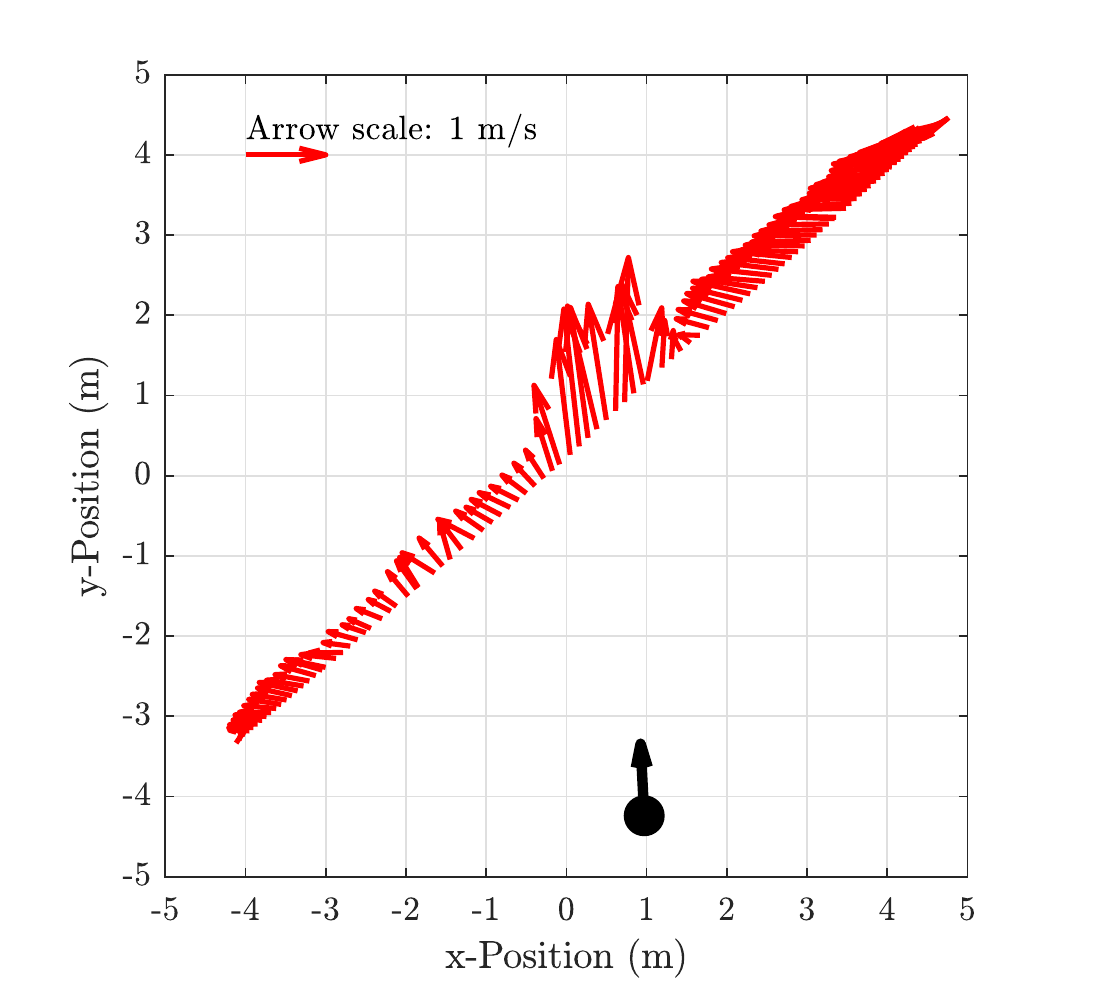}
    \vspace*{-.3in}
    \caption{In this plot the vehicle is flown in a straight line at high speed, from left to right, while a leaf blower (shown in black) aims at the middle of its trajectory. The red arrows indicate the intensity of the estimated wind speed.}
    \label{fig:wind_gust_detection}
\end{figure}

To demonstrate the ability to estimate wind gusts, we flew the vehicle in a straight line commanded by the trajectory generator outlined in \cref{subsec:implementation} along the diagonal of the flight space while a leaf blower was pointing approximately to the middle of this trajectory. \Cref{fig:wind_gust_detection} shows in red the estimated wind speed of the \ac{UKF} drawn at the 2D position where this value was produced, and in green the leaf blower pose obtained with the motion capture system. As expected, the wind speed is increased in the area affected by the leaf blower.

\begin{table}[t]
\vspace*{.15in}
\begin{center}
    \caption{RMS between LSTM and UKF in the estimation of the relative velocity of the robot on joystick dataset}
   		\label{tab:rmse_rel_vel} 
   		\begin{tabular}{ccccc}	
   			\hline
   			Method & RMS error $x$ & RMS error $y$ & RMS error $z$ & Unit\\
   			\hline
   			UKF	& 0.44 & 0.34 & 0.53 & m/s \\
   			LSTM & \textbf{0.38} & \textbf{0.31} & \textbf{0.28} & m/s \\
   		\end{tabular}
   	\end{center}
   	\vspace*{-.25in}
   \end{table}
\subsection{Simultaneous estimation of drag and interaction forces}
\begin{figure*}
\centering
\includegraphics[trim=0 0 0 0, clip, width=1\textwidth]{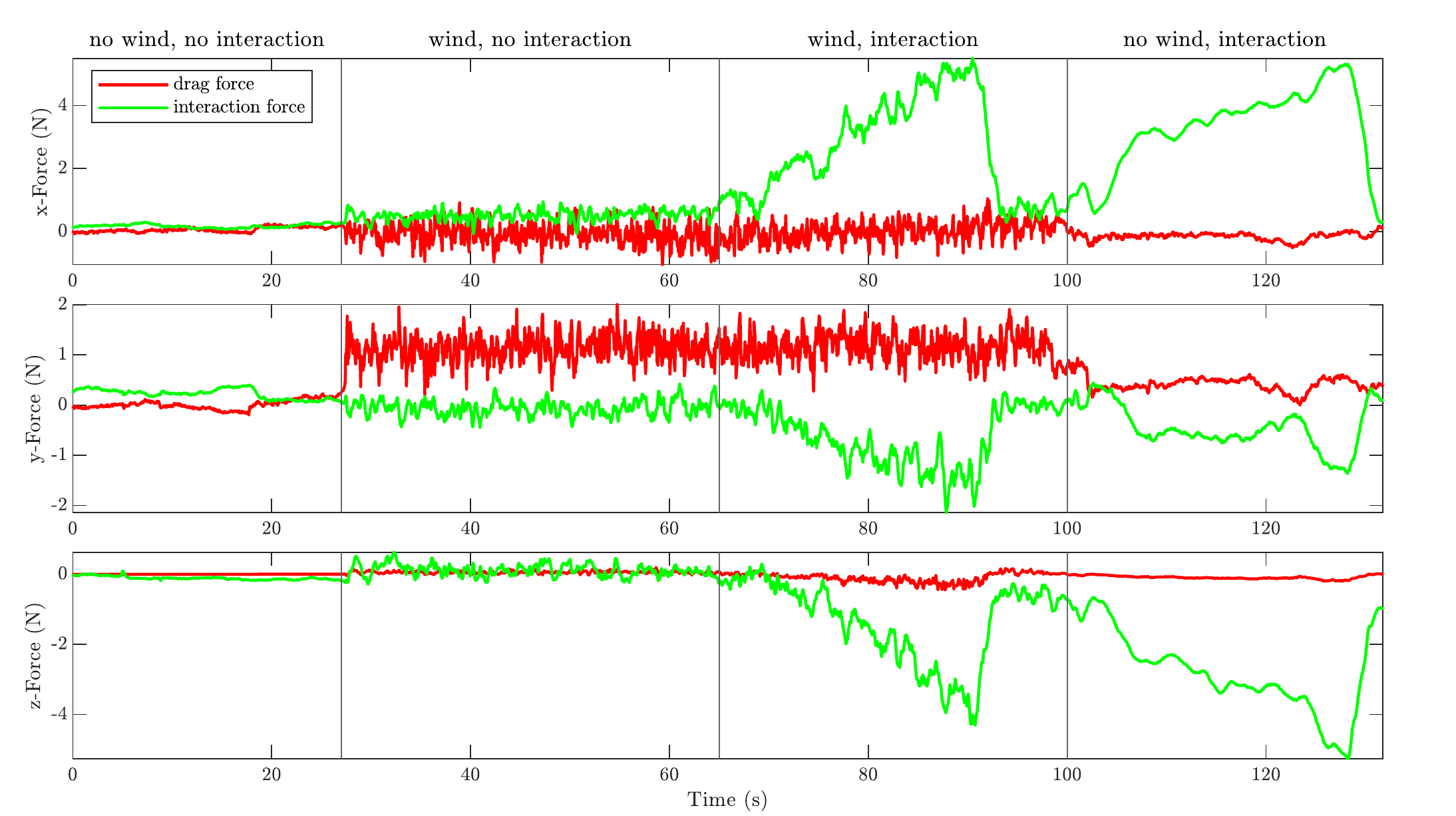}
\vspace*{-.3in}
\caption{Simultaneous estimation of drag and interaction force. Vertical bars separate the four phases of the experiment.}
\label{fig:simultaneous_drag_interaction}
\end{figure*}

Our approach can differentiate between drag and interaction forces, which is shown in the following experiments.
There are four main parts to the experiment: hovering with no external force, hovering in a wind field generated by three leaf blowers, simultaneously pulling the vehicle with a string attached to it while the vehicle is still immersed in the wind field, and turning off the leaf blowers so that there is only interaction force.
\Cref{fig:simultaneous_drag_interaction} shows the forces acting on the \ac{MAV} in world frame estimated by the \ac{UKF}: $\prescript{}{W}{\mathbf{f}}_{\text{drag}}$ and $\prescript{}{W}{\mathbf{f}}_{\text{touch}}$.
As expected, the drag force is close to zero when no wind is present even when the \ac{MAV} is pulled, and similarly the interaction force is approximately zero when the vehicle is not pulled even when the leaf blowers are acting on it.
Therefore, drag and interaction forces are differentiated correctly.
Note that the leaf blowers turn on quickly and thus the drag force resembles a step, while the interaction force was caused by manually pulling the \ac{MAV} with a string following approximately a ramp from 0 to 4 N as measured with a dynamometer.
The \ac{UKF} estimates $\prescript{}{W}{\mathbf{f}}_{\text{touch}}$ to about 6N, potentially due to inaccuracies of our external force ground truth measurement procedure and mis-calibration of the commanded throttle to thrust mapping. As for the wind speed generated by the leaf blowers, it has an average value of 3.6 m/s at the distance where the vehicle was flying. According to our model, a drag force of approximately 1.2 N as shown in \cref{fig:simultaneous_drag_interaction} should correspond to a wind speed of 3 m/s. The difference is due to the fact that the leaf blowers are not perfectly aimed to the \ac{MAV}, and the wind field that they generate is narrow.

\section{CONCLUSION}
\label{sec:conclusion}
We presented a model- and a learning-based approach to estimate the relative airflow, the drag force and the interaction force acting on a hexarotor using bio-inspired sensors. 
The results obtained in flight experiments show that our approach allows to accurately identify the relative airflow experienced by a multirotor in flight, and that we are able to detect wind gusts acting on the \ac{MAV}.  Via experimental results, we showed that the proposed deep-learning based strategy is more accurate than the model-based strategy, and does not require a significant amount of training data. The deep-learning based strategy, however, does not allow the flexibility to re-position or easily change sensors without having to re-train the network.
Additionally, we show that we can correctly distinguish between drag and interaction forces.
Future work includes leveraging our drag estimation results for improved trajectory tracking performance. We additionally plan to further evaluate our deep-learning based approach, and to compare different learning algorithms, data collection, and training strategies.

\section*{ACKNOWLEDGMENT}
This work was funded by the Air Force Office of Scientific Research MURI FA9550-19-1-0386 and by Ford Motor Company. The authors would like to thank Parker Lusk for his help in the system setup.

\balance

\bibliographystyle{IEEEtran}
\bibliography{bibliography.bib}

\begin{acronym}
\acro{MAV}{Micro Aerial Vehicle}
\acro{UKF}{Unscented Kalman filter}
\acro{EKF}{Extended Kalman filter}
\acro{LSTM}{Long Short-Term Memory}
\acro{RNN}{Recurrent Neural Network}
\acro{CoM}{center of mass}
\acro{USQUE}{Unscented Quaternion Estimator}
\acro{UT}{Unscented Transformation}
\acro{MSE}{Mean Squared Error}
\end{acronym}

\end{document}